# Identifying Phrasemes via Interlingual Association Measures – A Data-driven Approach on Dependency-parsed and Word-aligned Parallel Corpora

## 1. Introduction

It has been understood for a long time that the semantic content of a combination of two or more words often cannot be derived from the semantics of the single words, but that the use of one particular word imposes restrictions upon others (Firth 1957; Evert 2004, 15–17). The semantics is then either determined by the ruling word, e.g., in the case of light verb constructions (*attention* entails *pay* in *pay attention*), or by the entity of all participating words, e.g., in the case of idiomatic expressions or set phrases (*so to speak*).

Many names have been given to this phenomenon, each of which looking at it from a slightly different perspective: collocations, multiword expressions, phrasemes, idioms or formulaic sequences, to name just a few. The term lexical function (Wanner 1996; Mel'čuk 1998) stresses the aspect of one word being the value returned by a function applied to another word.[1] That lack of flexibility of the determined word, which does not contribute much – if anything – to the meaning of the composed expressions is what we make use of in the approach described in this paper.

In our work, we address the issue of phraseme identification in one language by searching for corresponding syntactic structures in parallel, word-aligned corpora. We exemplify our approach by retrieving and ranking support verb constructions[2] that consist of a verb and its direct object. The support verb's nature allows for correspondences that can be regarded as translations only in the context of the whole construction. A suitable translation of the support verb construction *pay attention* into German is *Aufmerksamkeit schenken* 'attention' + 'give as a present'/'make a gift'. While *attention* and *Aufmerksamkeit* embody the same semantic concept, *pay* and *schenken* can hardly be seen as good translations except for this particular case – and only in conjunction with their direct objects.[3]

This paper is structured as follows: In section 2, we give an overview of statistical association measures and their motivation. Section 3 explains the design of our corpus including the choice of corpus material and the annotation and alignment tasks that are required to allow for complex corpus queries such as the ones we use for phraseme identification. Interlingual association measures are introduced and discussed in section 4, which subsequently exemplifies their use by means of intralingual corpus queries for support verb constructions. In section 5, we discuss the application of our approach to other phenomena and present an outlook on possible enhancements.

---

1 The lexical function Oper$_1$, which specifies the verb used to perform the operation determined by the noun given as argument, would return *fare 'make'* or *porre 'put'* for the Italian noun *domanda*.

2 Support verb constructions are frequently seen as synonymous with light verb constructions which likewise places emphasis on the little semantic content that the verb brings along, but we want to allow for any verb to support the noun's semantics, i.e., to make it become performed (Mel'čuk 1998).

3 In fact, out of 179 aligned sentences where *pay* is aligned to *schenken* in our parallel corpus of English and German, 124 cases (69 %) correspond to *pay attention*/*Aufmerksamkeit schenken*. *Beachtung schenken* accounts for 45 cases of *pay attention* (25 %) and *Augenmerk schenken* for 4 cases (2 %). There is less variation in English.

## 2. Statistical Association Measures

Evert (2004; 2008) gives a profound overview of collocations including the differing concepts that other authors link to that concept. In this paper, we limit ourselves to the phraseological view of "lexically determined word combinations" (ibid.) and apply the statistical association measures used for ranking collocations by their "strongness" to word alignments obtained by statistical methods. Word alignments are relations between tokens in parallel sentences, i.e., sentences in two different languages that can be regarded as translations of each other.[4] Tiedemann (2011) deals with the topic of alignment on different levels and explains the concepts and algorithms used for obtaining them from raw parallel texts.

A statistical association measure is usually defined between two words, where word does not refer to a particular occurrence, i.e., a token, but to its prototypical form, the lemma. For a specific context, which may be a sentence, a span restricted to a particular number of tokens to the left and to the right, or a syntactic relationship, we count occurences of both words ($O_{11}$), just the first word but not the second one ($O_{12}$) and vice versa ($O_{21}$) as well as cases where none of the words are found ($O_{22}$). These observed frequencies form the basis of any of the association measures.[5] The sample size N is the sum of all the observed frequencies.

Given the sample size, the expected frequency $E_{11}$ is defined as the number of times one would expect to see both words together if the distribution of words concerning that relationship was random. The respective measure is interpreted as showing no association if both, the observed and the expected frequency, show roughly the same value. If the observed frequency is considerably higher than the expected frequency, the words are regarded as collocates, a higher ratio denoting a stronger collocation. The reverse holds for an observed frequency being considerably lower than the expected one.

Different association measures are known to have different shortcomings. For instance, O/E and pointwise mutual information (MI) which give the same ranking, but on different scales, favor rare word combinations. There is no single association measures known to always deliver best results. Especially for intralingual word association measures, no prior work has been done to our knowledge. The information theoretical foundation of, at least, some measures suggests that they should also be applicable to other relations such as word alignment.

---

[4] There is generally no restriction made on the direction of translation. One sentence can be the result of a translation process from the other language or both derive from a third source language.

[5] See (Evert 2004, pp. 35–40; Evert 2008, pp. 17–18).

## 3. Our Corpus

This section motivates our decisions for selecting the source material and further processing steps we carried out, that is, mostly levels of annotation and alignment on top of token sequences. This material forms the basis of empirical contrastive analysis of linguistic phenomena in the *Sparcling* project.[6] One of the principal interests in this project is the investigation of variable article use in English, which is where the combination of word alignment and dependency parsing shows its strengths as illustrated in Figure 1. We compare English with Finnish, French, German, Italian, Polish and Spanish.

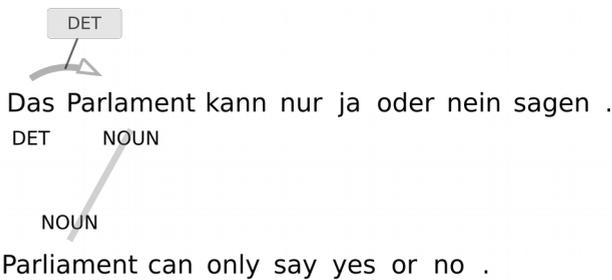

*Figure 1: In this example, the German noun phrase "das Parlament" translates to English article-less "Parliament". Both nouns are aligned.*

### 3.1 Source Material

Several multilingual corpora are freely available, but not as many as one would expect given today's state of globalization where translations are to be found everywhere: in furniture assembly instructions and food packaging, software localization and movie subtitles, not to mention the large number of translated books. Comparable corpora like Wikipedia[7] may serve as resources for particular tasks such as statistical machine translation (cf. Plamada and Volk 2013). For linguistic research, however, parallel corpora are preferred.

Östling (2015, pp. 5–6) gives an overview of large parallel corpora and compares them in terms of size and number of languages. Most of them are less suited for our kind of linguistic analysis, either because of being too small, not covering the languages required or, in the case of the subtitle collection *OpenSubtitles* (Tiedemann 2012), the text type. One of the remaining large parallel corpora that includes a variety of languages is *Europarl* (Koehn 2005) which comprises the debates of the European Parliament over 15 years (from 1996 to 2011). The debates are transcribed, which involves turning speech into grammatically correct text, and subsequently translated into all other official languages[8].

---

6   http://pub.cl.uzh.ch/purl/sparcling_project

7   https://www.wikipedia.org/

The derivation from speeches given at the plenary sessions qualifies Europarl for linguistic investigation rather than other corpora from the context of the European Union that cover the same quantity of languages but mostly contain legal texts (Steinberger et al. 2014). The Digital Corpus of the European Parliament (DCEP) by Hajlaoui et al. (2014) provides diverse text types, ranging from lists of agenda items and minutes of the parliamentary sessions to written and oral questions of members of the European Parliament and would presumably also serve as data source.

The European Parliament's sittings are subdivided into chapters which cover a particular topic. While some of those chapters show a formal character (votings for instance), others are close to discussions, including interruptions. In addition to contributions by particular speakers, which are transcribed and subsequently translated into all other official languages, members of the European Parliament are allowed to hand in written statements as individual members or in groups. That results in statements like Figure 2 which again broaden the linguistic scope of this corpus.

**de:** Deshalb mußten wir gegen diese Punkte stimmen.

**en:** It has therefore been necessary for us to vote against these paragraphs.

**it:** Ci siamo quindi visti costretti a votare contro questi punti.

*Figure 2: A sentence from a joint statement of three members.*

The downside of the Europarl corpus is that it comprises a multitude of errors, some of which have a serious negative impact on further processing steps. Tokenization errors may result in faulty annotation of the whole sentence, for instance. In (Graën, Batinic, and Volk 2014), we systematically classified the errors in the Europarl corpus and estimated their frequencies. We consequently corrected most of the traceable errors and published our revised version of the Europarl corpus under the name *Corrected & Structured Europarl Corpus (CoStEP)*.[9]

Sections with unrecoverable errors were dropped altogether. Furthermore, we dropped untranslated texts, i.e., speaker contributions that consisted of the same text in more than one language, and all texts that were only available in one language.

The CoStEP corpus is thus a smaller but cleaner version of Europarl. It comprises approximately 87 % of the original material. We are confident that the gain of quality on different levels compensates for the loss of a small amount of corpus material.

---

8   "Documents in less widely spoken languages are first translated into one of the three most commonly used relay languages (English, French or German) and then into other languages." (http://www.euparliament.eu/european-multilingualism; August 21st 2016)

9   The CoStEP corpus is available at http://pub.cl.uzh.ch/purl/costep.

Bartsch and Evert (2014) provide evidence that "larger corpora do not necessarily lead to better results" and "composition and cleanness of a corpus are more important than corpus size".

Beyond the raw texts, we aligned speaker turns and added meta information to the respective speakers from a list of members of the European Parliament obtained from the European Union.

## 3.2  Corpus Preparation

We extracted parallel text units, i.e., in our case, speaker turns from the sittings of the European Parliament, in English, Finnish, French, German, Italian, Polish and Spanish from the CoStEP corpus. Altogether, 136,298 speaker turns are available in all of these languages, except for Polish, which covers only 43,458 turns (32 %).

All texts were tokenized with our in-house tokenizer Cutter, which we equipped with token identification rules tailored to the parliamentary domain. Language-specific rules based on word forms, lemmas and part-of-speech tags allowed us to not only identify ordinary sentence boundaries but also sentence segment boundaries, which, in addition, split parts of sentences by colon or semicolon and make up more than 6 % of the segments in our corpus.

For tagging and lemmatization, we used the TreeTagger (Schmid 1994) with the language models available from the TreeTagger's web page.[10] To increase tagging accuracy for words unknown to the TreeTagger language model, we had to extend the tagging lexicons, especially the German one, with lemmas and part-of-speech tags for frequent words. Moreover, we used the word alignment information between all the languages (see below) to disambiguate lemmas for tokens where TreeTagger provided multiple lemmatization options. Our lemma disambiguation approach works like the one described by Volk, Amrhein, et al. (2016), except that we take evidence from all aligned tokens in any language into account for the disambiguation decision.

For German verbs with separable prefixes, the verb lemma is usually given without its prefix since the tagging algorithm cannot handle complex syntactic structures. In the sample sentence *"Ich drücke mich in Französisch aus, damit ich direkter verstanden werden kann." (I am speaking French in order to make myself understood more directly.)*, the word *drücke* is lemmatized to *drücken 'push/press'*, whereas the the correct lemma would be *ausdrücken 'express'* with the prefix *aus* being placed at the end of the clause. In principle, these corrections could be made by means of the syntactic structure of the sentence. However, deriving a correct syntactic structure depends in particular on correct part-of-speech tags, which is why this step is typically based on tagging. We describe the algorithm for reattaching German verb prefixes in (Volk, Clematide, et al. 2016).

On the sentence segments identified (about 1.7 million per language), we performed pairwise sentence alignment with hunalign (Varga et al. 2005) and, based on that, word alignment with GIZA++ (Och and Ney 2003; Gao and Vogel 2008) and the Berkeley Aligner (Liang, Taskar, and Klein 2006). Word alignment was performed on the types of all tokens and on lemmas of content words.[11] For the latter, we mapped the language models' individual tagsets to the universal tag set defined by Petrov, Das, and McDonald (2012) and defined content words to be those tokens that are tagged as either nouns, verbs, adjectives or adverbs.

---

10 http://www.cis.uni-muenchen.de/~schmid/tools/TreeTagger/#Linux

We used the *MaltParser* (Nivre, Hall, and Nilsson 2006) to derive syntactical dependency relations in German, English and Italian. For parsing English, we had to map several part-of-speech tags beforehand as tagger and parser were trained on different versions of the Penn Treebank tagset. Since there are no pre-trained models available for the MaltParser, we built our own. To do so, we obtained the *Italian Stanford Dependency Treebank (ISDT)*[12] from the *evaluation campaign of Natural Language Processing and Speech tools for Italian (EVALITA)*[13] and replaced tags and lemmas with the respective fields returned by the Italian TreeTagger model. In addition to the tagset used by the Italian TreeTagger model, we also added universal part-of-speech tags from the mapping mentioned previously. We continued by training a parsing model on the modified treebank using the *MaltOptimizer* (Ballesteros and Nivre 2012). Figure 2 depicts the different levels of linguistic information.

This method adds tagging errors to the gold data and thus leads to a performance loss compared with a model trained on the unmodified treebank, which can only be applied to input data with the same tagset. Furthermore, the lack of morphological information in the TreeTagger output is also likely to decrease parsing accuracy.

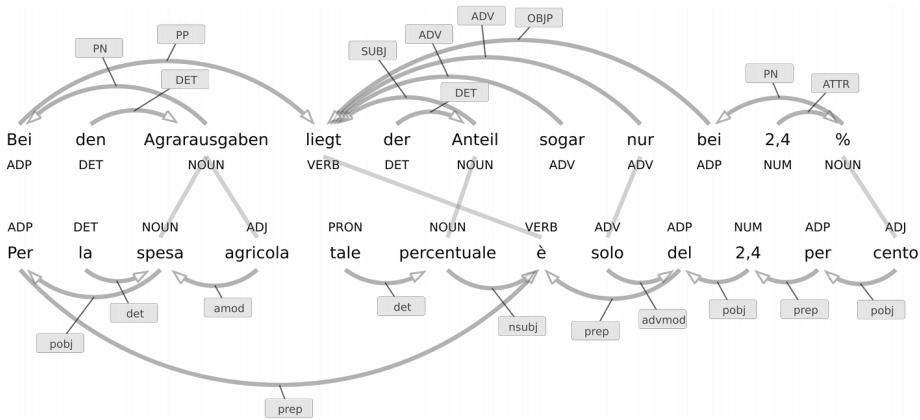

*Figure 2: Two corresponding sample sentences in German and Italian. Four levels of linguistic information are shown: Token boundaries, universal part-of-Speech tags (small caps), syntactic dependency relations with labels (arcs) and word alignment on content words (straight lines). Sentence-final periods have been left out.*

---

11  We used the word form instead if no lemma was provided.
12  http://medialab.di.unipi.it/wiki/ISDT
13  http://www.evalita.it/

## 4. Interlingual Association Measures

The main idea behind any association measure of words in a corpus is that their ability to construct a joint semantic concept when appearing together is reflected by their frequencies in the corpus. Revisiting the example from the introduction, we can say that *pay* when keeping company of *attention* – to follow Firth's (1957, p. 11) famous quote – merely modifies the semantics of *attention,* but does not contribute its own regular semantics.[14]

We can calculate association measures similar to the ones we use for a monolingual corpus between the aligned words of corresponding sentences in two languages. These interlingual association measures give an account of the interlingual company a word keeps, i.e., which two words are seen together more frequently than random word alignment would entail. Similar to intralingual association scores, a higher interlingual association score represents a stronger translation preference. Those interlingual associations are assumed to be symmetric here.[15]

For those words representing semantic concepts, the noun in case of support verb constructions, we thus expect a high interlingual association score, whereas the association score of the supporting verbs is expected to be low in most cases. The raw frequencies of those verbs often vary. Our sample comprises 1209 occurrences of *pay* and only 319 occurences of *schenken.* It is thus not surprising that *pay* translates to *zahlen 'pay'* or *bezahlen 'pay for something'* (60 % of the cases) four times more often than to *schenken* (15 %) while *pay* is the most frequent translation of *schenken* (56 % of the cases). The fact that the expression *Aufmerksamkeit schenken* outnumbers any other use of *schenken* (*give* and *devote* only account for 24 % and 12 %, respectively) is most likely due to the parliamentary domain.

### 4.1 Interlingual Corpus Queries

Our corpus comprises word alignment for all language pairs and syntactic dependency parses for English, German and Italian, as described in section 1.3.2. Based on both of these data layers, we query for all tuples $(t_1^1, t_1^2, t_2^1, t_2^2, d_1, d_2)$ such that token $t_1^1$ (the verb) of language $L_1$ is aligned with token $t_2^1$ of language $L_2$ and likewise $t_1^2$ (the noun) with $t_2^2$ and syntactic relations with dependency labels $d_1$ and $d_2$ are defined from $t_1^2$ to $t_1^1$ and from $t_2^2$ to $t_2^1$ respectively. In total, 22 million of those complex structures are found in about 120 million tokens that these three languages

---

[14] The Macmillan online dictionary (https://www.macmillandictionary.com/dictionary/british/pay_1) gives "to give money in order to buy something" as first definition, Merriam-Webster (http://www.merriam-webster.com/dictionary/pay) reads "to make due return to for services rendered or property delivered" (as of August 21st 2016).

[15] In the sample used in this section, *Aufmerksamkeit* translates to *attention* in 97 % of the cases, while the other way round, *attention* translates to *Aufmerksamkeit* in only 91 %. Other frequent translations for *attention* that together with *Aufmerksamkeit* cover 98 % of the cases are *Beachtung* and *Augenmerk.*

comprise together. For each language, we have approximately 40 million tokens in our corpus.

We subsequently defined a subset of tuples $(l_1^1, l_1^2, l_2^1, l_2^2)$ for each language pair by mapping all tokens $t_L^n$ to their respective lemma $l_L^n$ and filtering for dependency relations that denote direct object relationship in the respective dependency label sets. The last element denotes the absolute frequency that we observe for a particular combination of lemmas. In total, we get 325,000 tuples for the language pair German/Italian, 376,000 tuples for German/English, and 605,000 tuples for English/Italian.[16] Aggregating unique combinations of lemmas, we get a list of distinct tuples for each language pair together with their absolute frequency. Table 1 shows the top five of these lists.

| rank | German | | English | | Italian | | count |
|---|---|---|---|---|---|---|---|
| 1 | spielen | Rolle | play | role | | | 2095 |
| 2 | haben | Recht | have | right | | | 1181 |
| 3 | unterstützen | Bericht | support | report | | | 1084 |
| 4 | finden | Lösung | find | solution | | | 983 |
| 5 | unterstützen | Vorschlag | support | proposal | | | 799 |
| 1 | spielen | Rolle | | | svolgere | ruolo | 1726 |
| 2 | haben | Recht | | | avere | diritto | 1241 |
| 3 | lösen | Problem | | | risolvere | problema | 1088 |
| 4 | finden | Lösung | | | trovare | soluzione | 845 |
| 5 | treffen | Entscheidung | | | prendere | decisione | 716 |
| 1 | | | take | account | tenere | conto | 2522 |
| 2 | | | play | role | svolgere | ruolo | 2432 |
| 3 | | | thank | rapporteur | ringraziare | relatore | 2142 |
| 4 | | | have | right | avere | diritto | 1890 |
| 5 | | | draw | attention | richiamare | attenzione | 1320 |

*Table 1: Most frequent combination of aligned verb and direct object composition for each language pair.*

The most common corresponding verb/object collocations are domain-specific (e.g., *support (a) report/proposal*) and not particularly idiomatic. To allow for a reasonable extraction of relevant tuples, we calculate association measures for the syntactic relations as well as for the alignments. Since we had no prior expectation towards which measure would give the best results, we performed the calculation for different measures, namely the observed/expected ratio (O/E), pointwise mutual information (MI), local-MI, z-score, t-score and simple likelihood (simple-ll). Evert (2008, section 4) explains them in detail.

---

16 The lower numbers for combinations with German suggest that these are due to structural differences between German and both, English and Italian. Especially compounds are known to be a challenge for word alignment as they increase the number of word types and hence lead to lower probabilities for the word alignment model.

The sets of tuples and association measures for each language can be obtained at http://pub.cl.uzh.ch/purl/interlingual_association_measures.

## 4.2   Identifying Support Verb Constructions

The characteristics of support verb construction are – as a result of lacking semantics – that given the direct object $t_1^2$ in language $L_1$ and the corresponding support verb construction consisting of tokens $t_2^1$ and $t_2^2$ in a second language $L_2$,[17] the verb $t_1^1$ that fits into that constellation cannot be derived, but has to be learned. That applies to statistical language models on training data as well as to human language learners of language $L_1$. The most common translation of the verb $t_2^1$ will presumably fail if a support verb construction is concerned. Translating, for instance, the English expression take *precedence (over)* to German, *Vorrang* will be a good fit for *precedence.* A literal translation of *take* will lead to *nehmen,* which is no combination a German dictionary would licence.[18] The other way around, *have precedence,* the literal translation of *Vorrang haben,* is licenced by dictionaries though dispreferred.

In order to approach the objective to extract lists of the most relevant support verb constructions for each language pair, we experimented with different configurations of the association measures that we had calculated. Finding a formula that reliably calculates a good ranking based on the association measures is complicated by the fact that there is no limit in how high numbers for the associations can get. The respective numbers are also not comparable; the distribution of intralingual association scores differs between languages. To address this incompatibility, we applied a cumulative percentile ranking (cpr) to all the scores, mapping them to values within the range from 0 to 1.

We stated earlier that we expect the nouns to be matching translations – hence a high alignment association score –, whereas the support verbs in the majority of cases will not match – hence a low alignment association score. Accordingly, we get an initial ranking of the tuples by calculating the ratio of both alignment scores for a given measure $am_x$ :

$$q(x,\delta) \;=\; \frac{\delta + cpr\left(am_x(l_1^2, l_2^2)\right)}{\delta + cpr\left(am_x(l_2^1, l_1^1)\right)}$$

---

[17] Via alignment of $t_1^2$ and $t_2^2$ and the direct object relation of $t_2^2$ and $t_2^1$.

[18] We will still find examples for the literal translation of the whole expression when searching the internet. This is can possibly be explained with bad machine translation and the inability of search engines to recognize a source as such, though efforts have been made to distinguish original from machine translated texts. Moreover, search engines also find and index online communication of language learners who initially resort to literal translations.

This fraction can result in very high values for low association scores on the verb alignment. The delta (δ) introduced to the ratio puts a limit on which values it can get. A delta of 1, for instance, limits the possible values to the range from 0.5 to 2.

To also take the intralingual association scores into account, we multiplied the ratio with their weighted arithmetic mean:

$$r(x,y,\alpha,\beta,\delta) \;=\; (\alpha \times am_y(l_1^1, l_1^2) \;+\; \beta \times am_y(l_2^1, l_2^2)) \;\times\; q(x,\delta)$$

This gives for local-MI as *x*, O/E as y, and *α, β* and *δ* set to 1 the lists shown in Table 2 with an additional frequency filter of $f \geq 2$ applied. Without filtering, we get good high-ranked matches such as *Arbeitslosigkeit bekämpfen/reduce unemployment* or *Ausnahme zulassen/make exception* for German/English, but we must not overlook the risk of annotation or alignment errors.[19] A filter for at least two congruent matches seems a reasonable remedy since for the same error two occur twice, both annotation and aligmnent layers need to coincide twice which reduces the probability of error to the multiplied error rates of all four layers.

| rank | German | | English | | Italian | | count |
|---|---|---|---|---|---|---|---|
| 1 | annehmen | Gestalt | take | shape | | | 39 |
| 2 | darstellen | Präzedenzfall | set | precedent | | | 10 |
| 3 | bekämpfen | Armut | reduce | poverty | | | 4 |
| 4 | schaffen | Präzedenzfall | set | precedent | | | 78 |
| 5 | haben | Vorrang | take | precedence | | | 47 |
| 1 | schaffen | Abhilfe | | | porre | rimedio | 36 |
| 2 | schaffen | Präzedenzfall | | | costituire | precedente | 23 |
| 3 | gewinnen | Oberhand | | | prendere | sopravvento | 8 |
| 4 | machen | Mühe | | | prendere | briga | 9 |
| 5 | schaffen | Klarheit | | | fare | chiarezza | 6 |
| 1 | | | take | look | dare | occhiata | 21 |
| 2 | | | take | precedence | dare | precedenza | 4 |
| 3 | | | send | condolence | esprimere | condoglianza | 5 |
| 4 | | | take | precedence | avere | precedenza | 92 |
| 5 | | | have | illusion | fare | illusione | 20 |

*Table 2: Highest ranked combinations of aligned verb and direct object composition according to r(local-MI, O/E, 1, 1, 1)*

Lowering the value of δ results in tuples with a lower absolute frequency being ranked higher. Other association measures put a different weight on the lightness of the verb and result in them being ranked higher or lower.

---

19 Evert (2008) states that "[t]heoretical considerations suggest a minimal threshold of f ≥ 3 or f ≥ 5, but higher thresholds often lead to even better results in practice."

## 5. Discussion and Outlook

We presented a novel approach of identifying phrasemes on parallel corpora. Dependency parsing and word alignments are required levels of linguistic information on top of the sentence aligned parallel corpus, both of which methods are based on statistical models and, therefore, error-prone. Querying the corpus for a complex constellation such as the aligned verb and direct object relationships shown in section 4.2, we minimize the risk of repeated errors, which allows us to choose the lowest possible frequency threshold. Only constellations occuring once need to be discarded, and those still hold interesting cases as a manual examination showed.

One of the limitations of our approach is that we commit ourselves to a particular syntactic structure that we search for in both languages of the respective parallel corpus. That way, we are unable to find translations with diverging structures. Support verb constructions in one language, can often be translated using a single verb, e.g., German/Italian *Fortschritte machen/avanzare 'make progress'*. Moreover, the combination of verb and noun lemma sometimes is not enough to indicate an expression. Aggregating over the syntactic context of the expression *prendere briga* from Table 2, one could probably derive the whole expression *prendersi la briga (di fare qc.) 'to take pains (with sth.)'*.

The association measures we used for our experiments do not consider the asymmetric nature of support verb constructions. There are several – more complex – association measures that take into account that in a collocation of two words, one might be the result of having chosen the other one, like it is the case for support verb construction, but also a multitude of other phenomena in natural language, for which those measures might give even better evidence.

Another interesting attempt is the extraction of parallel syntactic constellations from more than two languages, e.g., from German, English and Italian, simultaneously. This approach would presumably increase the quality of corresponding structures identified, but lower their number considerably. Use cases, such as the extraction of similar expressions in two languages that differ from a third one, might be of interest for language learners who already have a good command of two languages and want to spot difficulties in the third one. Take, for instance, the support verb constructions *Entscheidung treffen/take decision/prendere decisione*, where the semantics of the verbs *take* and *prendere* coincide, but differ from German *treffen 'strike/meet'*.

## 6. Acknowledgements

This research was supported by the Swiss National Science Foundation under grant 105215_146781/1 through the project "SPARCLING – Large-scale Annotation and Alignment of Parallel Corpora for the Investigation of Linguistic Variation". The author likes to thank his colleagues Noah Bubenhofer, Simon Clematide and Samuel Läubli for discussions and proof-reading. Special thanks are owed to everyone who contributed to our corpus: Chiara Baffelli, Dolores Batinic, Christof Bless, Simon Clematide, Stéphanie Lehner, Mathias Müller, Martin Volk, Sara Wick, and Ventsislav Zhechev.

References

Ballesteros, M. and J. Nivre (2012). "MaltOptimizer: an optimization tool for MaltParser". In: *Proceedings of the Demonstrations at the 13th Conference of the European Chapter of the Association for Computational Linguistics (EACL)*.
Bartsch, S. and S. Evert (2014). "Towards a Firthian notion of collocation". In: *Network Strategies, Access Structures and Automatic Extraction of Lexicographical Information. 2nd Work Report of the Academic Network Internet Lexicography, OPAL–Online publizierte Arbeiten zur Linguistik*.
Evert, S. (2004). "The Statistics of Word Cooccurrences: Word Pairs and Collocations". PhD thesis. Universität Stuttgart.
Evert, S. (2008). "Corpora and collocations". In: *Corpus linguistics. An international handbook 2*. Ed. by A. Lüdeling and M. Kytö.
Firth, J. (1957). "Modes of Meaning". In: *Papers in Linguistics 1934–1951*. Oxford University Press.
Gao, Q. and S. Vogel (2008). "Parallel implementations of word alignment tool". In: *Software Engineering, Testing, and Quality Assurance for Natural Language Processing*.
Graën, J., D. Batinic, and M. Volk (2014). "Cleaning the Europarl Corpus for Linguistic Applications". In: *Proceedings of the Conference on Natural Language Processing (KONVENS)*.
Hajlaoui, N., D. Kolovratnik, J. Väyrynen, R. Steinberger and D. Varga (2014). "DCEP-Digital Corpus of the European Parliament". In: *Proceedings of the 9th International Conference on Language Resources and Evaluation (LREC)*.
Koehn, P. (2005). "Europarl: A parallel corpus for statistical machine translation". In: *Machine Translation Summit*. Vol. 5.
Liang, P., B. Taskar, and D. Klein (2006). "Alignment by agreement". In: *Proceedings of the main conference on Human Language Technology Conference of the North American Chapter of the Association of Computational Linguistics*.
Mel'čuk, I. (1998). "Collocations and Lexical Functions". In: *Phraseology. Theory, Analysis, and Applications*. Ed. by A. P. Cowie.
Nivre, J., J. Hall, and J. Nilsson (2006). "Maltparser: A data-driven parser-generator for dependency parsing". In: *Proceedings of the 5th International Conference on Language Resources and Evaluation (LREC)*. Vol. 6.
Och, F. J. and H. Ney (2003). "A Systematic Comparison of Various Statistical Alignment Models". In: *Computational linguistics*. Vol. 29.
Petrov, S., D. Das, and R. McDonald (2012). "A Universal Part-of-Speech Tagset". In: *Proceedings of the 8th International Conference on Language Resources and Evaluation (LREC)*.

Plamada, M. and M. Volk (2013). "Mining for Domain-specific Parallel Text from Wikipedia". In: *Proceedings of the Sixth Workshop on Building and Using Comparable Corpora*.

Schmid, H. (1994). "Probabilistic part-of-speech tagging using decision trees". In: *Proceedings of International Conference on New Methods in Natural Language Processing (NeMLaP)*. Vol. 12.

Steinberger, R., M. Ebrahim, A. Poulis, M. Carrasco-Benitez, P. Schlüter, M. Przybyszewski, and S. Gilbro (2014). "An overview of the European Union's highly multilingual parallel corpora". In: *Language resources and evaluation*.

Tiedemann, J. (2011). "Bitext Alignment". In: *Synthesis Lectures on Human Language Technologies*.

Tiedemann, J. (2012). "Parallel Data, Tools and Interfaces in OPUS". In: *Proceedings of the 8th International Conference on Language Resources and Evaluation (LREC)*.

Varga, D., L. Németh, P. Halácsy, A. Kornai, V. Trón, and V. Nagy (2005). "Parallel corpora for medium density languages". In: *Proceedings of the Recent Advances in Natural Language Processing (RANLP)*.

Volk, M., C. Amrhein, N. Aepli, M. Müller, and P. Ströbel (2016). "Building a Parallel Corpus on the World's Oldest Banking Magazine". In: *Proceedings of the Conference on Natural Language Processing (KONVENS)*.

Volk, M., S. Clematide, J. Graën, and P. Ströbel (2016). "Bi-particle Adverbs, PoS-Tagging and the Recognition of German Separable Prefix Verbs". In: *Proceedings of the Conference on Natural Language Processing (KONVENS)*.

Wanner, L. (1996). *Lexical functions in lexicography and natural language processing*. Vol. 31. John Benjamins Publishing.